\documentclass{article}
\pdfoutput=1
\usepackage{spconf,amsmath,graphicx}

\usepackage{algorithmic}
\usepackage{algorithm}

\usepackage{caption}

\usepackage{subfigure}

\title{Improving Spectral Clustering Using Spectrum-Preserving Node Aggregation}

\name{Yongyu Wang}
\address{University of Southern California, Los Angeles, CA 90089, USA}

\begin{document}
%\ninept
%
\maketitle
\begin{abstract}
Spectral clustering is an important graph clustering technique. However, it suffers from a scalability problem due to the involved computationally expensive eigen-decomposition procedure and is highly sensitive to noisy nodes in the graph. In this work we solve the two problems simultaneously by using  spectrum-preserving node aggregation to generate a nearly-noiseless concise representation of the data set. Specifically, we create a small set of spectral-representative pseudo-nodes based on spectral analysis. Then, standard spectral clustering algorithm is performed on the smaller node set. Finally, we map the clusters of pseudo nodes back to find the cluster memberships of the data points in the original data set. The proposed framework has a nearly-linear time complexity. Meanwhile, the clustering accuracy can be significantly improved. The experimental results show dramatically improved clustering performance when compared with state-of-the-art methods.

%For instance, our method achieves $18\%$ and $15\%$ accuracy gain on MNIST and USPS benchmarks over the standard spectral clustering method, respectively.

\end{abstract}
\section{Introduction}
\label{sec:intro}

Clustering is one of the most fundamental machine learning problems \cite{nie2009spectral}. It aims to assign data samples in a data set into different clusters in such a way that samples in the same cluster are more similar compared to those in different clusters.

%Our strategy is based on observation that concise pattern can help to improve clustering pattern \cite{wang2021high}

 In the past decades, many clustering techniques have been proposed. Among them, spectral clustering has drawn considerable attention and has been widely applied to computer vision, image processing and speech processing \cite{bach2004learning,wan1998new}. Although spectral clustering has superior performance, the involved eigen-decomposition procedure has a time complexity of \textbf{$O(N^3)$}, where $N$ is the number of samples in the data set. The high computational cost can easily become an obstacle in many large-scale applications \cite{liu2013large,chen2011large}. To solve this problem, considerable effort has been devoted in research related to approximate spectral clustering. \cite{yan2009fast} proposed a k-means based hierachical spectral clustering framework (KASP); Inspired by sparse coding theory, \cite{chen2011large} proposed a landmark-based representation method (LSC);  \cite{fowlkes2004spectral} leveraged the Nystr\"om method to approximate the affinity matrix. However, none of these methods can preserve the spectrum of the original graphs, thus result in a significant degradation of clustering accuracy. For example, for Covtype data set, the clustering accuracy drops more than $26\% $ when using the LSC method, more than $21\% $when using the {KASP} framework and more than $15\% $when using the Nystr\"om method.
 
\cite{zhao2021towards} attempted to apply a graph coarsening method for dividing the graph topology into random number of partitions. However, its claims about spectral clustering have the following problems:
 
1) Methodologically, the algorithm flow in \cite{zhao2021towards} only partitions very few pseudo aggregated nodes. Without mapping the pseudo clusters to original real nodes via mapping operator, \cite{zhao2021towards} can only provide an auxiliary intermediate result for data clustering.

2) Partitioning pure graph topologies into small parts and finding clusters of data samples are two different tasks. Clustering is the task of grouping similar samples into same clusters. According to the definition of spectral clustering \cite{von2007tutorial}, the input of spectral clustering is similarity matrix of data samples calculated from their feature vectors. However, in \cite{zhao2021towards}'s experiment about spectral clustering, there is no data sample but only pure graph topologies are used as input. Without "data" points represented by features, there will be no "similarity" among nodes. 

3) To convincingly evaluate the clustering performance, each input sample should have a ground truth label (cluster membership) so that we can check the correctness of the clustering algorithm's result \cite{chen2011large,liu2013large}. However, in \cite{zhao2021towards}, clustering correctness cannot be measured because there is no label. So the experiments in \cite{zhao2021towards} cannot demonstrate the effectiveness of spectrum-preserving node aggregation on spectral clustering.

4) In clustering, the number of clusters are meaningful or naturally formed based on the density. For example, handwritten digits tasks have 10 clusters corresponding to digits 0-9, ModelNet40 data set \footnote{https://modelnet.cs.princeton.edu/} has 40 clusters corresponding to  its 40 kinds of 3D objects (tables, planes, chairs, ...). However, \cite{zhao2021towards} chose a random number 30 as the number of subsets for all the graphs. This is not make sense for spectral clustering experiments and evaluations.

In contrast to \cite{zhao2021towards}, in this paper, we integrate the spectrum-preserving node aggregation into the general approximate spectral clustering framework \cite{yan2009fast}  and construct the correspondence table based on the mapping operators. Then we use the table to map the clusters of aggregated nodes to original samples in the data set to find cluster membership of real data points. Based on this complete framework, we use real-world data sets to demonstrate the effectiveness of the  aggregation scheme for data clustering. The major contributions of this work has been summarized as follows:\\

\begin{enumerate}
  \item In this work, we proposed a complete framework for improving spectral clustering by integrating spectral methods into the general approximate spectral clustering framework \cite{yan2009fast} and demonstrate its effectiveness via real-world data sets. \\
  \item In contrast to other spectral clustering acceleration methods based on representative nodes such as KASP, Nystrom and LSC, our method uses the spectral-representative nodes which can directly preserve the spectrum of the graph Laplacian which will be key to clustering accuracy. As a result, for very large data set such as \textbf{Covtype} which has $581,012$ instances, only our method can accelerate clustering without loss of accuracy. All the other representation nodes based methods have $15\%$ to $26\%$ accuracy degradation. \\
  
  \item Our method enable to construct a nearly-noiseless manifold representation of the data set and dramatically clustering accuracy improvements can be achieved. For example, by removing more than $99.84\%$ of nodes for the MNIST data set, its noisy nodes are also removed together. The clustering accuracy improves from $64.20\%$ to $82.94\%$.\\
  
\end{enumerate} \
 
\section{Preliminaries}\label{sect:preliminaries}

k-means is the most fundamental clustering method \cite{von2007tutorial}. It discovers $k$ clusters by minimizing the following objective function:

\begin{equation}\label{eqn:kmeansObj}
F={\sum\limits_{i = 1}^n \sum\limits_{j = 1}^k \eta_{ij} {\|\textbf{x}_i-\boldsymbol{\mu}_j\|^{}_{2}}},
\end{equation}

where:

\begin{equation}\label{formula_deg}
\eta_{ij}=\begin{cases}
1 & \text{ if } \textbf{x}_i \in C_j\\
0 & \text{otherwise },
\end{cases}
\end{equation}

and $\boldsymbol{\mu}_j$ is the centroid of the cluster $C_j$.

\begin{figure}[!h] \centering 
\subfigure[k-means of the two moons data set]{ \includegraphics[width=1.2in,height=0.8in]{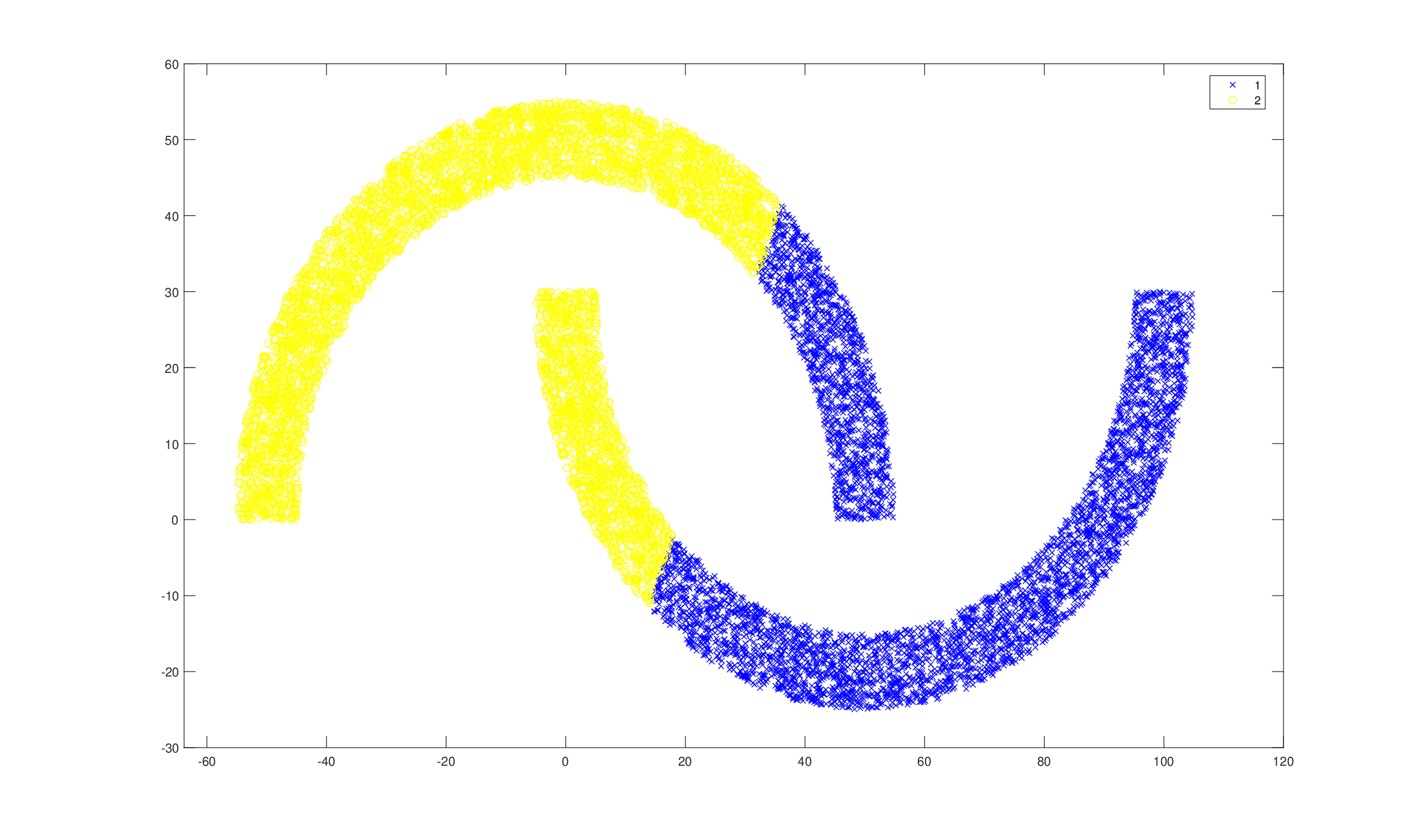} } \subfigure[Spectral clustering of the two moons data set ]{ \includegraphics[width=1.2in,height=0.8in]{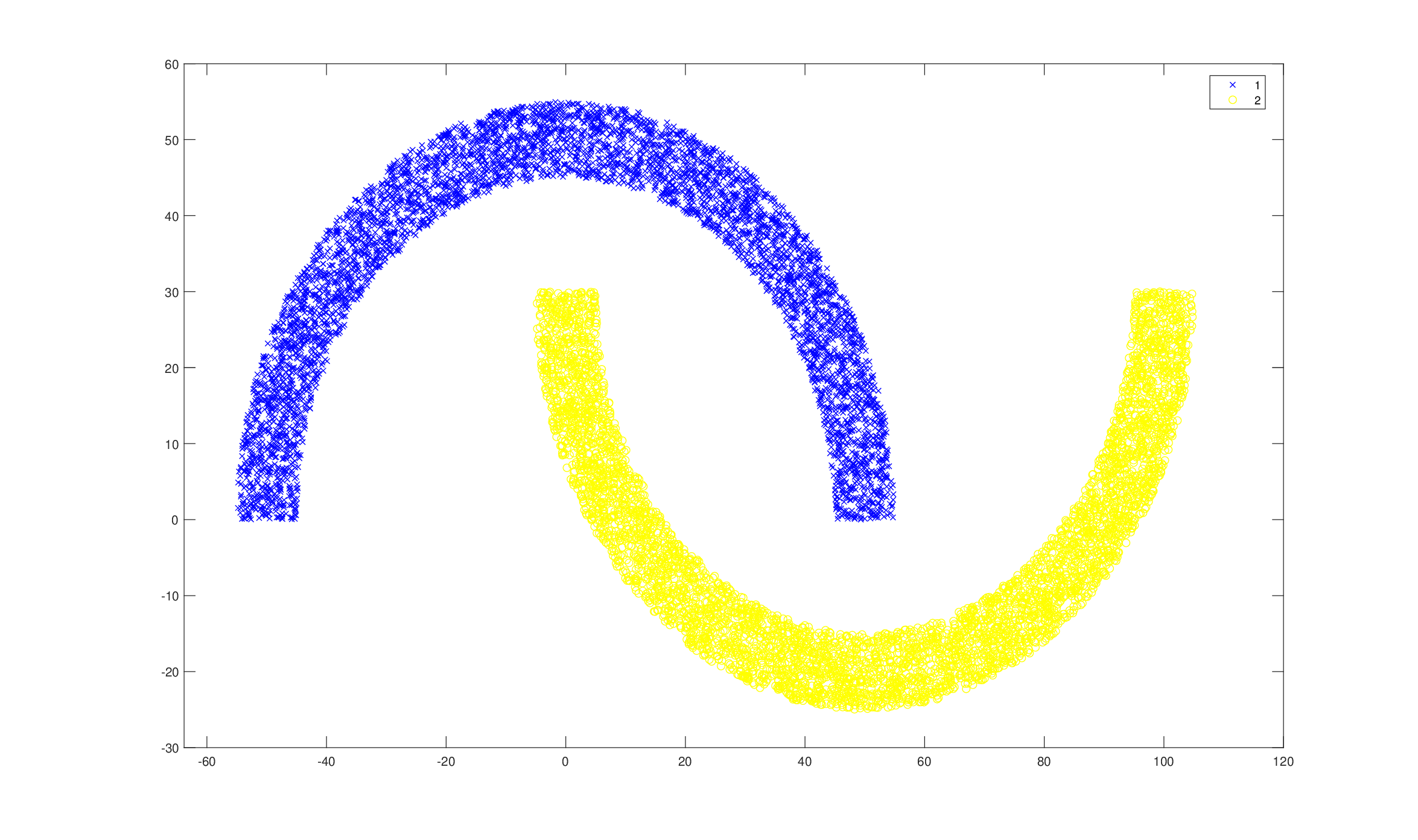} } \subfigure[k-means of the two circles data set]{ \includegraphics[width=1.2in,height=0.8in]{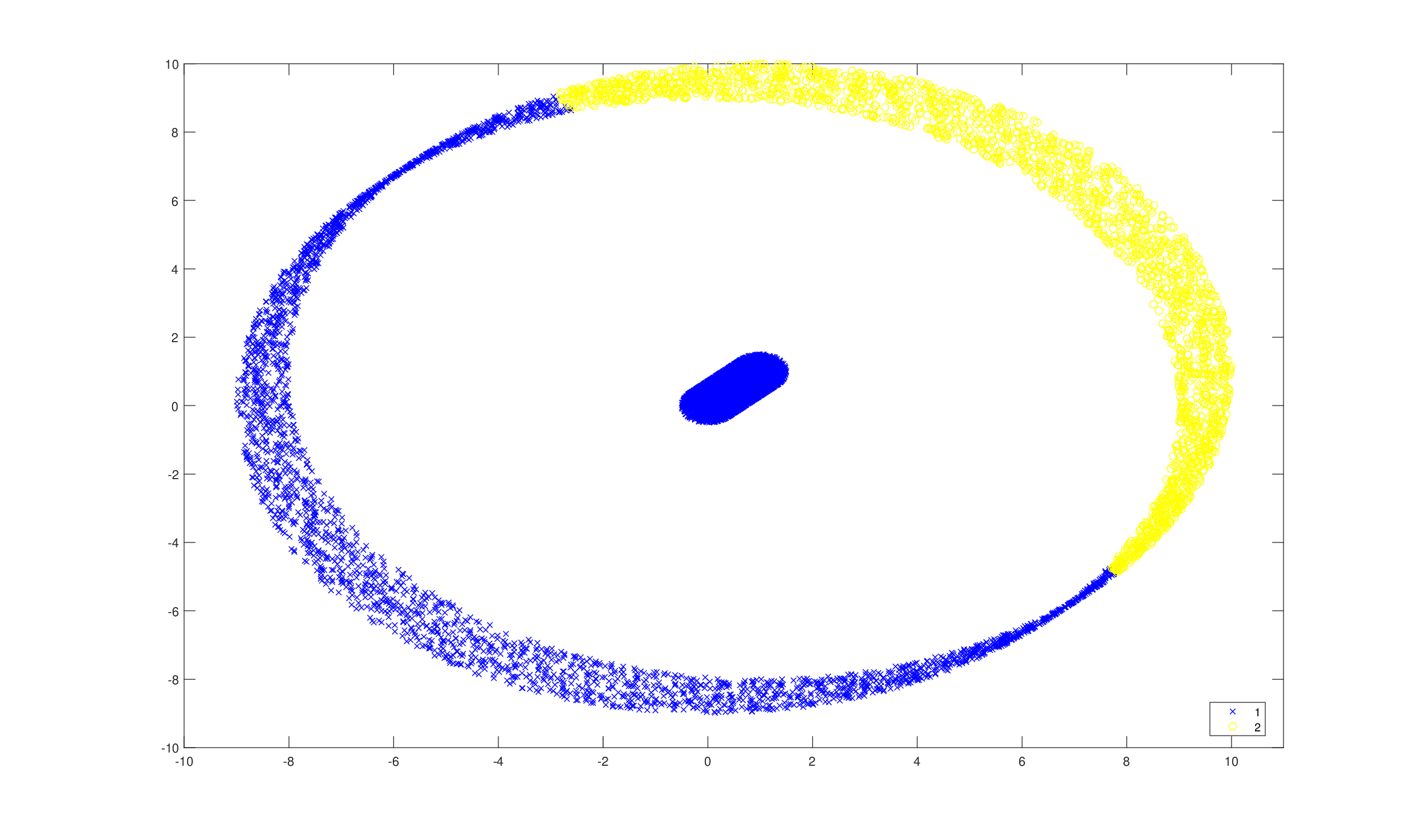} } \subfigure[Spectral clustering of the two circles data set]{ \includegraphics[width=1.2in,height=0.8in]{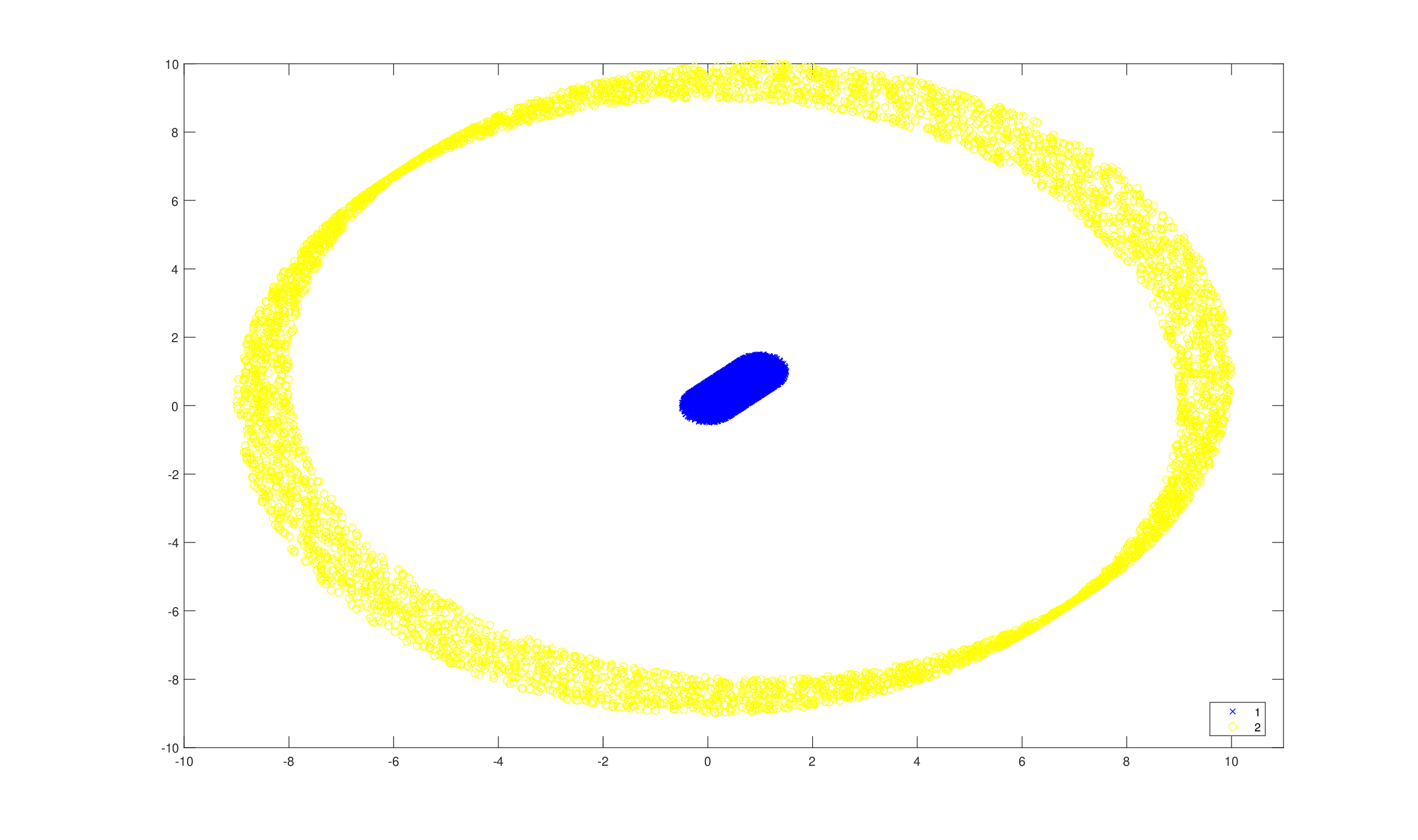} } 
\caption{Performance of k-means and spectral clustering} \label{fig:nonconvexshape} 
\end{figure}

However, it often fails to handle complicated geometric shapes, such as non-convex shapes \cite{nie2019k}. In contrast, spectral clustering is good at detecting non-convex and linearly non-separable patterns \cite{ng2002spectral}. Spectral clustering includes the following three steps: 1) construct a Laplacian matrix $L_G$ according to the similarities between data points; 2) embed nodes into $k$-dimensional space using the first $k$ nontrivial eigenvectors of $L_G$; 3) apply k-means to partition the embedded data points into $k$ clusters. As shown in Figure \ref{fig:nonconvexshape}, in the two moons data set, there are two slightly entangled non-convex shapes, where each cluster corresponds to a moon. In the two circles data set, the samples are arranged in two concentric circles, where each cluster corresponds to a circle. k-means gives incorrect clustering results for both data sets while spectral clustering performs an ideal clustering. Because k-means only cares about the distance while spectral clustering cares about the connectivity between nodes. The connectivity information is embedded in the spectrum of the underlying graph so that it is critical to preserve the spectrum when manipulating graphs \cite{wang2021high}.

 \section{Methods}
 
 \subsection{Algorithmic Framework}
\begin{figure}[!h] \centering 
\subfigure[The adjacency graph corresponding to the original node set of 9298 nodes]{ \includegraphics[width=1.9in,height=1in]{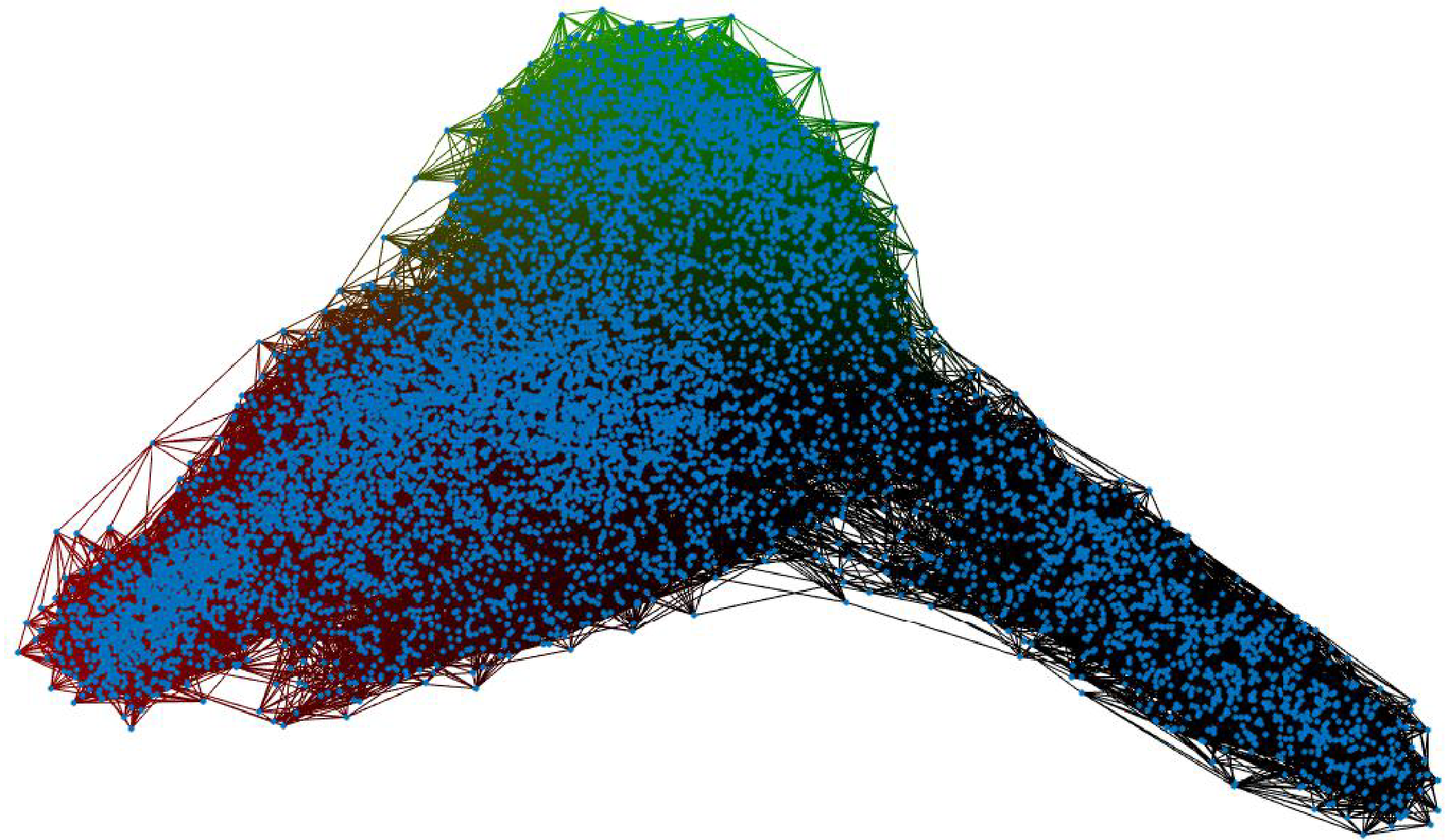} } \subfigure[The adjacency graph corresponding to the $50X$ reduced node set of 138 pseudo-nodes ]{ \includegraphics[width=1.9in,height=1in]{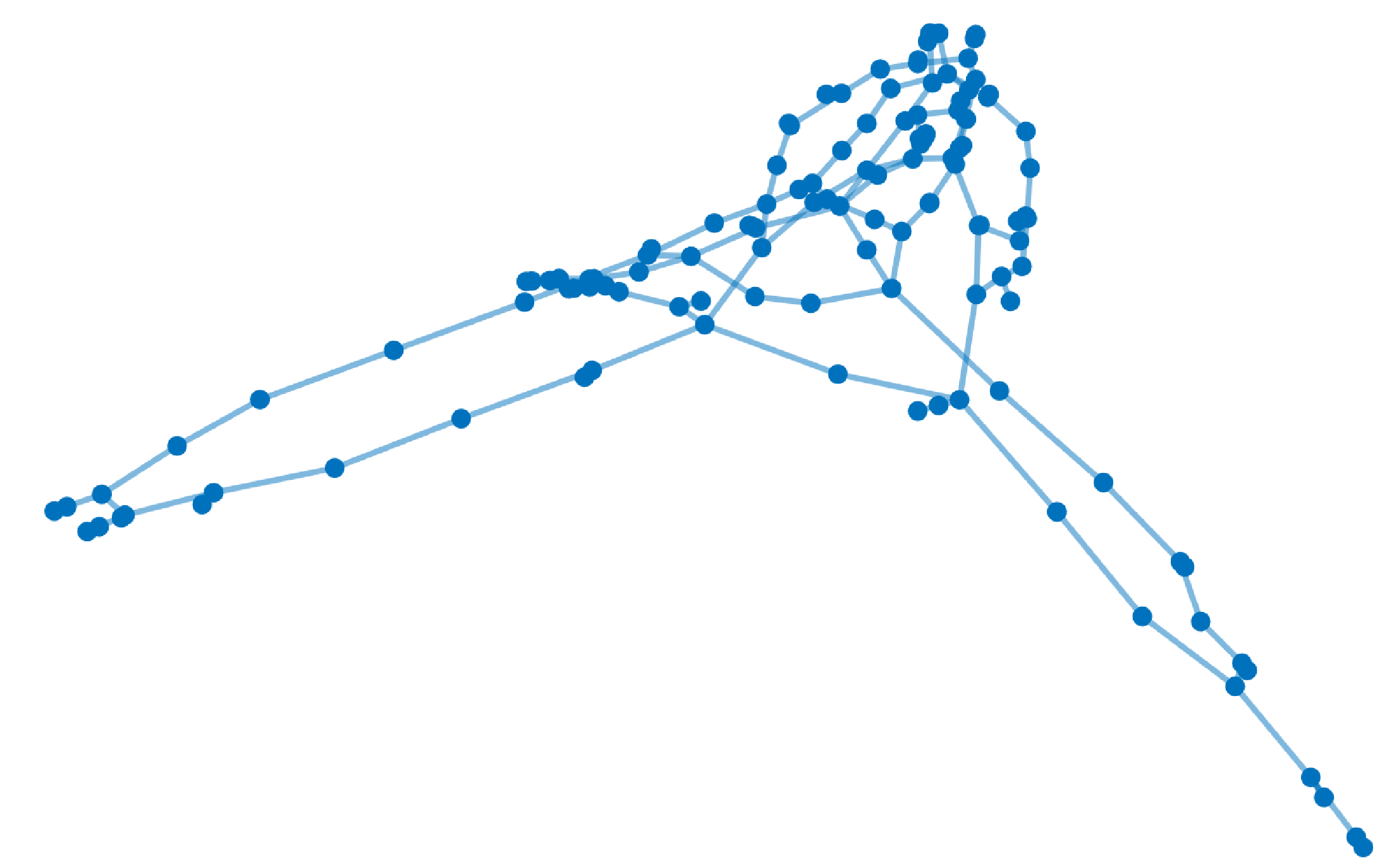} }  
\caption{Visualization of node aggregation results of USPS data set} \label{fig:compare1} 
\end{figure}

We first construct a standard k-NN graph $G$. Then, based on the node proximity measure proposed in \cite{livne2012lean}, the spectral similarity of two nodes $p$ and $q$ can be calculated as:
\begin{equation}\label{eqn:agg}
Sim_{p,q} =\frac{{|(\mathbf{X}_p, \mathbf{X}_q)|}^2}{(\mathbf{X}_p, \mathbf{X}_p)(\mathbf{X}_q, \mathbf{X}_q)},  \mbox{   } (\mathbf{X}_p,\mathbf{X}_q) := \sum_{k=1}^{K}{\left(\mathbf{x}_p^{(k)} \cdot \mathbf{x}_q^{(k)}\right)},
\end{equation}
where $\mathbf{X} := (\mathbf{x}^{(1)}, \dots, \mathbf{x}{^{(K)}})$ is obtained by applying Gauss-Seidel relaxation to solve ${L}_{G} \mathbf{x}^{(i)}=0$ for $i=1, \dots, K$, starting with $K$ random vectors. We aggregate the nodes with high spectral similarity to form a few spectral-representative nodes to reduce the graph size while preserving the spectrum of the original graph. By aggregating nodes based on spectral similarity, the reduced graph can best represent the original data set in the sense of minimizing the structural distortion \cite{wang2021high}, as shown in Figure \ref{fig:compare1}.

Then the standard spectral clustering is performed on the aggregated nodes. Spectral clustering algorithm needs to perform a  very computational expensive eigen-decompistion procedure with $O(n^3)$ time complexity to calculate the first $k$ non-trivial eigenvectors of the graph Laplacian, where $n$ is number of nodes. By using spectral-representative nodes to represent the global structure of the original data set, number of nodes $n$  can be reduced and thus leads to speedup of eigen-decomposition. If the desired size of node set is not reached, the nodes can be further aggregated layer by layer by utilizing a multi-level aggregation scheme \cite{zhao2021towards}: given the graph Laplacian ${L}_G$ and the mapping operators from the finest level $1$ to the coarsest level $r$, ${L}_R=\mathbf{H}_G^R {L}_G \mathbf{H}_R^G$, where $\mathbf{H}_G^R = \mathbf{H}_1^2 \mathbf{H}_2^3 \cdots \mathbf{H}_{r-1}^r$ and $\mathbf{H}_R^G = \mathbf{H}_2^1 \mathbf{H}_3^2 \cdots \mathbf{H}_r^{r-1}$

However, the goal of data clustering is to find cluster membership of original samples in the data set rather than the clusters of the pseudo aggregated nodes. In this paper, we propose to map the cluster membership back from the pseudo aggregated nodes to the original real data points. Specifically, we build a correspondence table using the mapping operators of each layer to associate each original sample to its aggregated nodes in the final layer. For each original data points, we assigning it to the cluster as its corresponding aggregated node. The complete algorithm flow has been shown in Algorithm \ref{alg:kaspnew}.

\begin{algorithm}[!h]
\small { \caption{ spectrum-preserving node aggregation-based high-performance spectral clustering framework} \label{alg:kaspnew}
\textbf{Input:} A data set $D$ with $N$ samples $x_1,...,x_N \in {R}^{d}$, number of clusters $k$.\\
\textbf{Output:} Clusters $C_1$,...,$C_k$.\\

\begin{algorithmic}[1]
    \STATE Construct a k-nearest neighbor (kNN) graph $G$ from the input data ; \\
    \STATE Compute the adjacency matrix $A_G$ and diagonal matrix $D_G$ of graph $G$  ; \\
    \STATE Compute the Laplacian matrix $L_G$=$D_G$-$A_G$;\\
    \STATE Perform spectrum-preserving node aggregation to obtain the reduced graph Laplacian $L_r$; \\
    \STATE Build a correspondence table to associate each node with its corresponding aggregated node;\\

    \STATE Compute the eigenvectors $u_1$,...,$u_k$ that correspond to the bottom $k$ nonzero eigenvalues of $L_r$;\\
    \STATE Construct $U \in {R}^{|V_{L_r}| \times k}$, with the $k$ eigenvectors stored as column vectors;\\
    \STATE Perform k-means algorithm to partition the rows of $U$ into $k$ clusters.\\
   
    \STATE Retrieve the cluster membership of each $x_i$ by assigning it to the cluster as its corresponding aggregated node and return the result.\\
    
\end{algorithmic}
}
\end{algorithm}

\subsection{Nearly noiseless representation}
 
One drawback of spectral clustering is that it is highly sensitive to noisy nodes in the graph, which limits the capability to accurately discover clusters: if a few noisy nodes form a narrow bridge between two densely connected communities, the algorithm tends to report the two communities plus the noisy nodes as a single community. 

Relying on the very strong capability of spectrum-preserving pseudo nodes in preserving the spectrum of the graph Laplacian and global manifold of the data set, almost all the original nodes including the noisy nodes can be removed together. For example, for the MNIST data set with $70,000$ nodes, only $109$ spectral-representative nodes are enough for representing the global structure of the data set, which means $99.84\%$ of nodes can be removed so that the noise nodes won't exist in the aggregated graph. As a result, the clustering accuracy improves from $64.20\%$ to $82.94\%$ for the MNIST data set.

Even though other approximate spectral clustering methods can also significantly reduced to node set size with a small amount of representative nodes, however, without using spectral analysis to choose spectrally-critical nodes,  their representative nodes are not able to truthfully encode the spectrum of original graph Laplacian. So their clustering accuracy gains from denoising are counteracted with the loss of spectrum information.

 \subsection{Algorithm Complexity}

The computational complexity of spectrum-preserving node reduction is $O(|E_G|log(|V|))$, where  $|E_G|$ is the number of edges in the original graph and $|V|$ is the number of nodes. The computational complexity of reduced standard spectral clustering is $O(P^3)$ where $P$ is the number of aggregated nodes. The complexity of cluster membership retrieving is $O(N)$, where $N$ is the number of samples in the original data set.

\section{Experiment}\label{sect:experiments}
Experiments are performed using MATLAB R2020b running on a Laptop. The reported results are averaged over $10$ runs.

%Node reduction scheme implementation is available at \footnote{https://github.com/cornell-zhang/GraphZoom/tree/master/mat\_coarsen}. 

\subsection{Experiment Setup}
One mid-sized data set (USPS), one large data set (MNIST) and one very large data set (Covtype) are used in our experiments. They can be downloaded from the UCI machine learning repository \footnote{https://archive.ics.uci.edu/ml/}  and LibSVM Data\footnote{https://www.csie.ntu.edu.tw/~cjlin/libsvmtools/datasets/}: \textbf{USPS} includes   $9,298$ images of USPS hand written digits with $256$ attributes;
\textbf{MNIST} is a data set from Yann LeCun's website\footnote{  http://yann.lecun.com/exdb/mnist/}, which includes  $70,000$ images with each of them represented by $784$ attributes; \textbf{Covtype} includes  $581,012$ instances for predicting forest cover type  from cartographic variables and each instance with $54$ attributes is from one of seven classes. 

%The statistics of these data sets are shown in Table~\ref{table:benchmark}.

%\begin{table}
%\begin{center}
%\normalsize\addtolength{\tabcolsep}{-2.5pt} \centering
%\caption{Statistics of the data sets.}
%\begin{tabular}{ |c|c|c|c|c|c|  }
% \hline   Data set        &   Size   &  Dimensions    &  Classes \\
 
% \hline   PenDigits          &  7,494  &  16   &    10           \\
% \hline   USPS     &  9,298  &  256   &    10          \\
% \hline   MNIST          &  70,000  &  784   &    10           \\
% \hline   Covtype         &  581,012  &  54   &    7           \\
% \hline
%\end{tabular}\label{table:benchmark}
%\end{center}
%\end{table}

We compare the proposed method against both the baseline and the state-of-the-art fast spectral clustering methods including:
(1) the standard spectral clustering algorithm \cite{von2007tutorial},
(2) the Nystr\"om method \cite{fowlkes2004spectral},
(3) the Landmark-based spectral clustering (LSC) method that uses random sampling for landmark selection \cite{chen2011large}, and
(4) the KASP method using k-means for centroids selection \cite{yan2009fast}. For fair comparison, we use the same parameter setting in \cite{chen2011large} for compared algorithms: the number of sampled points in Nystr\"om method ( or the number of landmarks in LSC, or the number of centroids in KASP ) is set to 500.

\subsection{Evaluation Metrics}

Clustering accuracy ($ACC$) is the most widely used measurement for clustering quality \cite{chen2011large,liu2013large}. It is defined as follows:
\begin{equation}\label{eqn:scale}
ACC= \frac{\sum\limits_{i = 1}^N  {\delta {(y_i,map(c_i))}}}{{N}},
\end{equation}
where $N$ is the number of data samples, $y_i$ is the ground-truth label, and $c_i$ is the label generated by the algorithm. $\delta (x,y)$ is a delta function defined as: $\delta (x,y)$=1 for $x=y$, and $\delta (x,y)$=0,  otherwise. $map(\bullet)$ is a permutation function which can be realized using the Hungarian algorithm \cite{papadimitriou1982combinatorial}. A higher value  of $ACC$ indicates better  clustering quality.

\vspace{-0.1in}
\subsection{Experimental Results}

\begin{table*}[!h]

\begin{center}

\caption{Spectral clustering accuracy (\%)}  
\scalebox{0.9}{
\begin{tabular}{ |c|c|c|c|c|c|c|c|c|c|c|c|c|c|c|c}

 \hline Data Set&Standard SC& Nystr\"om & KASP &LSC&Ours($5X$)&Ours($10X$)&Ours($50X$)\\
 
 %\hline  PenDigits  &74.36&71.99 &71.56  &74.25&75.35&&&&\\
 \hline USPS  &64.31&69.31 &70.62  &66.28&70.65&72.25&81.39\\    
 \hline  MNIST  &64.20&55.86  &71.23 &58.94&70.99&72.42&82.18\\
 \hline Covtype &48.81&33.24  &27.56 &22.60&48.77&44.50&48.28\\
 
 \hline
\end{tabular}}\label{table:acc result}
\end{center}
\end{table*}

 \begin{table*}[!h]
\begin{center}
\centering
\caption{Runtime (seconds)}  
\scalebox{0.9}{
\begin{tabular}{ |c|c|c|c|c|c|c|c|c|c|c|c|c|c|c|c}

 \hline Data Set&Standard SC&Nystr\"om &KASP  &LSC&Ours($5X$)&Ours($10X$)&Ours($50X$)\\
 
 %\hline  PenDigits  &0.18&0.19&0.13 (0.49) &0.10&&&&&&\\
 \hline USPS  &0.72&0.29&0.16  &0.22 &0.21&0.18&0.17\\    
 \hline  MNIST &252.59&0.95 &0.18 &0.49&0.59&0.35&0.21\\
 \hline Covtype &128.70&5.48 &0.70  &3.77&0.92&0.50&0.49\\
 
 \hline
\end{tabular}}\label{table:clustering time pc}
\end{center}
\end{table*}

\begin{table*}[!h]
\begin{center}
\centering
\caption{Graph complexity comparison}
\scalebox{0.9}{
\begin{tabular}{ |c|c|c|c|c|c|  }
 \hline  Data Set & $|V_{orig}|$  & $|V_{reduced}|($5X$)$ & $|V_{reduced}|($10X$)$& $|V_{reduced}|($50X$)$\\

 \hline   USPS     &9,298  &1,692  &767&138   \\
 \hline   MNIST    &70,000  &13,658  &6,368&1,182 \\
 \hline   Covtype   &581,012  &104,260  &44,188&8,192  \\
 \hline
\end{tabular}}\label{table:density}
\end{center}
\end{table*}

%Fig. \ref{fig:curve.eps} shows Fig. \ref{fig:nodescurve.eps} shows 

\begin{figure*}[!htbp]
\centering\includegraphics[scale=0.35]{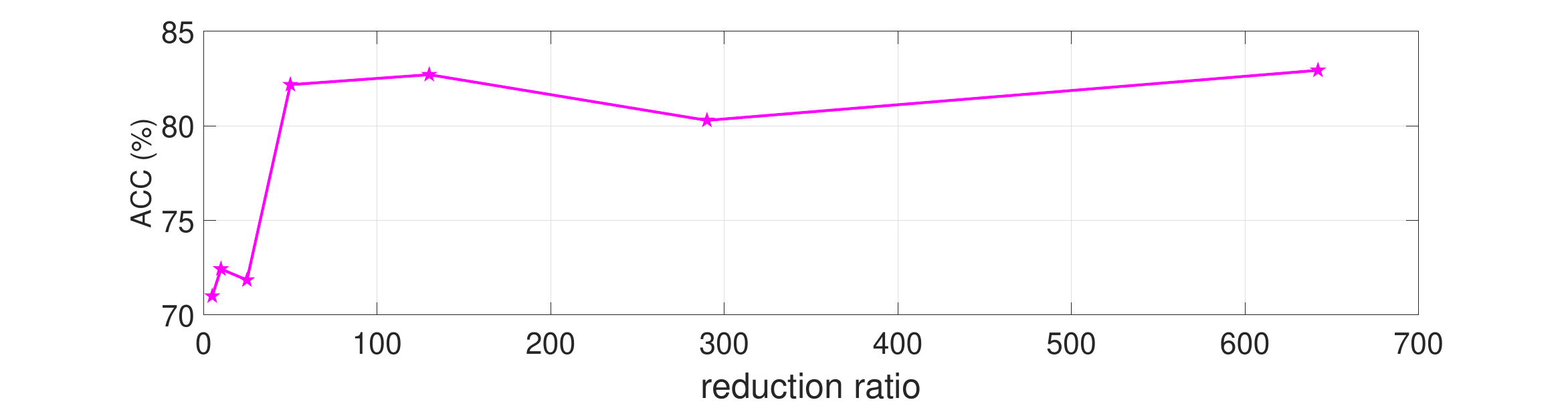}
\caption{ACC VS reduction ratio for the MNIST data set.\protect\label{fig:curve.eps}}
\end{figure*}

\begin{figure*}[!htbp]
\centering\includegraphics[scale=0.35]{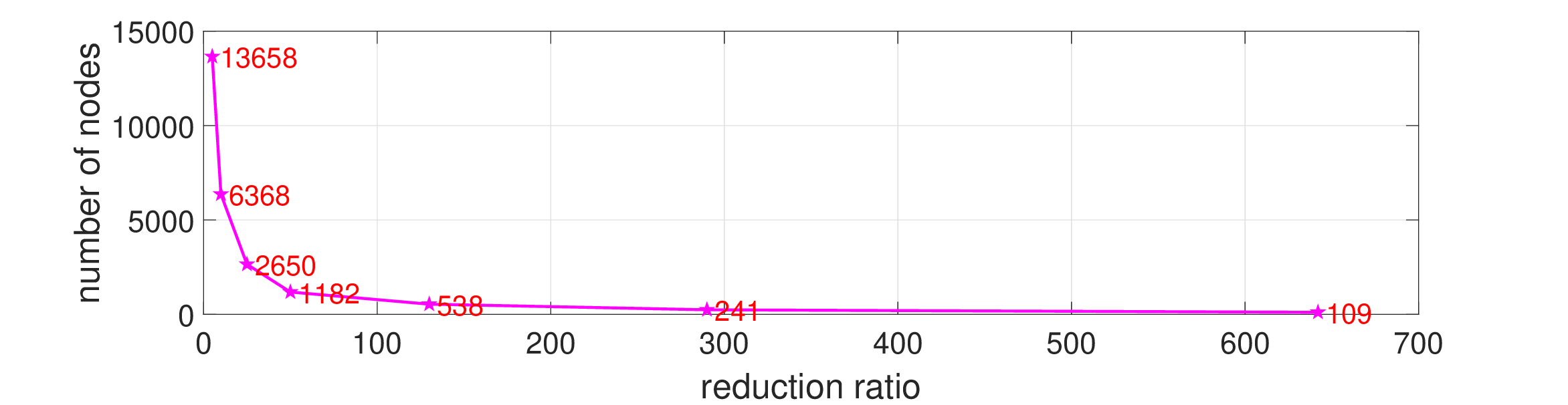}
\caption{Graph size VS reduction ratio for the MNIST data set.\protect\label{fig:nodescurve.eps}}
\end{figure*}

Table~\ref{table:acc result} shows the clustering accuracy results of compared methods and our method with node reduction ratios of $5X$, $10X$ and $50X$. The runtime of the eigen-decomposition and k-means steps are reported in Table~\ref{table:clustering time pc}. As observed, the proposed method can consistently lead to dramatic performance improvement, beating all competitors in clustering accuracy across all the three data sets: our method achieves more than $10\%$ accuracy gain on USPS, $11\%$ gain on MNIST and $15\%$ gain on Covtype  over the second-best methods; for the USPS and MNIST data sets, our method achieves over $17\%$ and $18\%$ accuracy gain over the standard spectral clustering method, respectively. The superior clustering results of our method clearly illustrate that spectrum-preserving concise representation can improve clustering accuracy by removing redundant or false relations among data samples.

As shown in Table~\ref{table:density}, the $50X$ reduced graphs generated by our framework have only $138$, $1182$ and $8192$ nodes for USPS, MNIST and Covtype data sets, respectively, thereby allowing much faster eigen-decompositions. As shown in Fig. \ref{fig:curve.eps} and Fig. \ref{fig:nodescurve.eps}, with increasing node reduction ratio, the proposed method consistently produces high clustering accuracy. For the very large data set Covtype, our method is the only one that enables us to accelerate spectral clustering algorithm without loss of clustering accuracy.

\section{Conclusion}\label{sect:conclusions}
In this paper, a complete framework for improving spectral clustering is proposed. We use real-world data sets to demonstrate the effectiveness of the method. Our method outperforms state-of-the-art methods by a large margin.

\vfill\pagebreak

\label{sec:refs}

\bibliographystyle{IEEEbib}
\bibliography{reference}

\end{document}